\documentclass{article}
\usepackage{iclr2026_conference,times}

\usepackage{amsmath,amsfonts,bm}

\def\eqref#1{equation~\ref{#1}}

\def\1{\bm{1}}

\DeclareMathAlphabet{\mathsfit}{\encodingdefault}{\sfdefault}{m}{sl}
\SetMathAlphabet{\mathsfit}{bold}{\encodingdefault}{\sfdefault}{bx}{n}

\usepackage{hyperref}
\hypersetup{pdftitle={Interventional Grounding Audits: Black-Box Premise-Dependency Tests for LLM Chain-of-Thought via Predicate Substitution}, pdfauthor={Hironao Nakamura}, hidelinks}
\usepackage{url}
\usepackage{booktabs}
\usepackage{amsmath}
\usepackage{amssymb}

\newcommand{\phiorig}{\phi_{\text{orig}}}
\newcommand{\phisem}{\phi_{\text{sem}}}
\newcommand{\phisur}{\phi_{\text{sur}}}

\newcommand{\artifacturl}{https://github.com/hironao-nakamura/interventional-grounding-audits/releases/tag/arxiv-v1.0}

\title{Interventional Grounding Audits: Black-Box Premise-Dependency Tests \\for LLM Chain-of-Thought via Predicate Substitution}

\author{Hironao Nakamura \\
Independent Researcher \\
}

\iclrfinalcopy
\begin{document}

\maketitle

\begin{abstract}
Large language models produce chain-of-thought (CoT) reasoning that appears logically sound yet may not genuinely depend on its stated premises. We introduce \emph{interventional grounding audits}, a \textbf{black-box, step-level test of premise dependency}: we intervene on a single premise by \textbf{substituting its target predicate with a fresh symbol}, re-run the model, and check whether each reasoning step's \textbf{normalized conclusion} (canonical predicate form) changes. We evaluate on \textbf{ProntoQA}, a synthetic multi-hop deductive reasoning benchmark with gold proof trees, where step-level premise dependencies are known. Applied to 50 ProntoQA problems with GPT-4o, our method achieves F1\,=\,0.806 on detecting proof-tree dependencies (F1\,=\,0.885 on predicate-determining dependencies; Recall\,=\,100\%), significantly outperforming a self-consistency baseline (F1\,=\,0.343; 95\% bootstrap CIs non-overlapping). We further identify that 66\% of correctly-solved problems contain at least one aligned step \emph{insensitive} to a direct proof-tree dependency under consistent substitution---all involving entity-introduction premises, a documented blind spot of the consistent-substitution evaluator---a ``right answer, wrong reasoning'' signal invisible to passive methods. All audit certificates, raw outputs, and reproduction scripts are available in a public GitHub repository, and we discuss scope limits beyond formal, parsable benchmarks.
\end{abstract}

\paragraph{Version note.}
This arXiv version corrects two issues in the accepted workshop artifact---a probe-generation issue (sequential predicate substitution could stack fresh-symbol prefixes, breaking chain coherence in some probes) and a CoT-to-proof step-alignment issue in the evaluation harness (model CoT steps are now matched to proof-tree steps by normalized conclusion, not by integer step number)---and re-collects all model outputs with the corrected pipeline (GPT-4o snapshot \texttt{gpt-4o-2024-08-06}, temperature 0). The accepted version's cross-model subject, Claude Sonnet~4 (\texttt{claude-sonnet-4-20250514}), was retired from the API before the rerun, so the cross-model analysis in Appendix~\ref{app:claude} uses the closest available successor, Claude Sonnet~4.5 (\texttt{claude-sonnet-4-5-20250929}), chosen to minimize model drift relative to the accepted version. Consequently, numbers differ from the accepted workshop version; all numbers in this version are recomputed on the unchanged 50-problem set and reproduced from the GitHub Release identified in Appendix~\ref{app:evidence}. The corrections strengthen rather than weaken the main results: F1 rises from 0.783 to 0.806 on all dependencies and from 0.835 to 0.885 on predicate-determining ones (Recall 97.4\%\,$\to$\,100\%), and the margin over the self-consistency baseline widens, with 95\% bootstrap CIs that remain non-overlapping. The most visible change is the RAWR rate (Section~\ref{sec:experiments}), which rises from 28\% to 66\% of correctly-solved problems under the corrected step alignment and now consists entirely of entity-introduction (structural) dependencies. Two secondary observations from the accepted version do not survive the corrections and are revised accordingly: the previously reported degradation of F1 with chain length disappears under corrected alignment (Appendix~\ref{app:chain}), and the cross-model ``meta-reasoning'' behavior is shown to be an artifact of the probe-generation issue rather than a model-specific strategy (Appendix~\ref{app:claude}). Appendix~\ref{app:evidence} records the full correction history.

\section{Introduction}
\label{sec:intro}

Chain-of-thought (CoT) reasoning can look logically sound while failing to genuinely depend on the premises it claims to use. CoT prompting enables large language models (LLMs) to solve multi-step reasoning problems by generating intermediate steps~\citep{wei2022chain}, but whether those steps are premise-dependent is often unclear. Consider an example from our experiments: GPT-4o solves a \textbf{ProntoQA} syllogistic problem correctly (a synthetic multi-hop deductive benchmark with gold proof trees, so each step's required premises are known), and self-consistency confirms this answer across five samples with 100\% agreement. Yet our method reveals an intermediate step that is \emph{insensitive} to one of its proof-tree premises---the model reaches the right answer while an audited step does not respond to intervention on its stated basis. This ``right answer, wrong reasoning'' (RAWR) signal appears in \textbf{66\% of correctly-solved problems} under the consistent-substitution evaluator (all such cases involve entity-introduction premises; Section~\ref{sec:experiments}).

Passive methods---self-consistency~\citep{wang2023selfconsistency}, attention analysis---observe outputs but never intervene on inputs. They cannot distinguish genuine logical dependency from correlation. Self-consistency achieves only F1\,=\,0.343 on our dependency detection benchmark, primarily because it cannot identify \emph{which} premises each step depends on (Precision\,=\,0.226).

We propose \emph{interventional grounding audits}: systematically substituting predicates in premises and observing whether each step's normalized conclusion changes. If replacing ``tumpus'' with an invented predicate ``glumpus'' in premise $P_j$ changes step $S_i$'s conclusion, $S_i$ genuinely depends on $P_j$. This approach draws on a two-layer grounding framework whose operational implementation achieves state-of-the-art on agent safety benchmarks~\citep{nakamura2026crosslayer,nakamura2026guard}, extended here from agent safety to reasoning integrity.

Our contributions: (1)~An interventional protocol detecting premise-level causal dependencies with F1\,=\,0.806 (0.885 on predicate-determining dependencies), significantly outperforming self-consistency (F1\,=\,0.343, non-overlapping CIs). (2)~Two substitution strategies---consistent and local---with cascade filtering that distinguish direct from transitive dependencies (best F1\,=\,0.819). (3)~A fully artifact-checkable evaluation: every certificate includes original and probed outputs with SHA256 checksums, verified by an automated validator.

\section{Method}
\label{sec:method}

\subsection{Problem Setup}

Given premises $\{P_1, \ldots, P_k\}$ and a question, an LLM generates a CoT with steps $\{S_1, \ldots, S_n\}$. A proof tree specifies ground-truth dependencies: for each $S_i$, a set $\text{deps}(S_i) \subseteq \{P_1, \ldots, P_k, S_1, \ldots, S_{i-1}\}$ of \textbf{direct} (immediate-parent) dependencies; we do not take the transitive closure. Our task: for each pair $(S_i, P_j)$, determine whether $S_i$ genuinely depends on $P_j$. Appendix~\ref{app:fp} discusses how 8\% of metric-counted false positives may be correct under a transitive definition. We model this through an \emph{observation layer} (raw LLM text) and a \emph{concept layer} (normalized propositions), intervening at the former and comparing at the latter.\footnote{This instantiates a formal grounding framework~\citep{nakamura2025observation,nakamura2026crosslayer}; see Appendix~\ref{app:formal}.}

\subsection{Interventional Protocol}

\textbf{Predicate substitution (semantic probe).}
To test whether $S_i$ depends on $P_j$, we replace the target predicate with an invented one (using a ``zq'' prefix). Two strategies: \emph{consistent substitution} replaces the predicate in \emph{all} premises, preserving chain coherence and detecting \emph{predicate-determining} dependencies; \emph{local substitution} replaces it \emph{only} in $P_j$, breaking the chain and detecting \emph{transitive} dependencies including structural premises. In ProntoQA, each premise introduces a unique predicate pair, so consistent predicate substitution is equivalent to premise-level intervention. In benchmarks with shared predicates across premises, this equivalence breaks down; local substitution addresses this case.

\textbf{Surface rephrasing (control probe).}
We rephrase $P_j$ without changing logical content (``All X are Y'' $\to$ ``Every X is a Y''). Output changes under surface rephrasing indicate surface sensitivity, not logical dependency.

\textbf{Normalized proposition extraction.}
Each step's conclusion is parsed into a canonical form---$\mathrm{is}(e, p)$ or $\mathrm{subtype}(p_1, p_2)$---absorbing irrelevant variation in phrasing.

\textbf{Five-value verdict.}
For each $(S_i, P_j)$, comparing normalized conclusions under original ($\phiorig$), semantic probe ($\phisem$), and surface probe ($\phisur$): \textsc{Grounded} ($\phiorig \neq \phisem$, $\phiorig = \phisur$); \textsc{Insensitive} (no change); \textsc{Input-Sensitive} (both change); \textsc{Unstable} (only surface changes); \textsc{Unparseable} (parse failure). An \textsc{Insensitive} step that explicitly cites $P_j$ constitutes a \textbf{misrepresentation}.

\subsection{Cascade Detection and Filtering}

Local substitution detects transitive dependencies but introduces cascade false positives: if local substitution on $P_j$ changes $S_i$, all downstream steps also change via propagation. Our cascade filter reclassifies $S_i$ as \textsc{Cascade} if $S_{i-1}$ is also \textsc{Grounded} w.r.t.\ the same $P_j$, recovering Precision (0.604\,$\to$\,0.756) while retaining Recall. This filter exploits the linear chain structure of ProntoQA; tree-structured proofs require generalization.

\subsection{Connection to Deployed Guard Architecture}

Our pipeline mirrors a deployed safety guard achieving state-of-the-art on OS-level agent safety~\citep{nakamura2026guard}: both use a two-stage \emph{assess}\,$\to$\,\emph{decide} pipeline with named decision rules, structured verdict dataclasses, and automated evidence validators. Same architecture, different domain.

\section{Experiments}
\label{sec:experiments}

\subsection{Setup}

We evaluate on ProntoQA~\citep{saparov2023language}, 50 synthetic syllogistic problems (3--5 hops, 4--7 premises, True/False balanced). Target model: GPT-4o (snapshot \texttt{gpt-4o-2024-08-06}, temperature\,=\,0). Dataset: 1,127 audit certificates, of which 1,031 are aligned to proof-tree steps for primary evaluation. Certificate-level parse rates were 95.3\% for original outputs, 92.9\% for semantic probes, 69.2\% for local probes, and 90.3\% for surface probes. For the combined protocol, 108/1,031 aligned certificates (10.5\%) were \textsc{Unparseable} and excluded from P/R/F1 computation. 17 CoT steps (in 8 problems) were unmatched to proof-tree conclusions---typically premise-restatement or unparseable steps---and excluded from primary metrics. Conversely, 9 proof-tree steps (11 premise dependencies) had no matching CoT step; these are counted as false negatives only in the lower-bound analysis in Section~\ref{sec:discussion}. No problem had an ambiguous alignment, so no problem-level exclusions were required. Appendix~\ref{app:evidence} reports the alignment policy, and all counts are reported in the released artifact repository. Metrics: P, R, F1 with 95\% bootstrap CIs ($B$\,=\,10{,}000, problem-level resampling). Two evaluation modes: $\text{F1}_\text{full}$ (all proof-tree dependencies) and $\text{F1}_\text{pred}$ (predicate-determining only).

\subsection{Main Results}

\begin{table}[t]
\caption{Main results (GPT-4o, ProntoQA 50 problems). Bootstrap CIs confirm significance.}
\label{tab:main}
\vspace{0.3em}
\centering
\small
\begin{tabular}{@{}lcccc@{}}
\toprule
Method & P & R & F1 & 95\% CI \\
\midrule
Self-Consistency & 0.226 & 0.715 & 0.343 & [0.317, 0.371] \\
String-diff & 0.801 & 0.774 & 0.787 & --- \\
A consistent (full) & 0.794 & 0.820 & 0.806 & [0.760, 0.852] \\
A+A$'$+cascade (full) & 0.756 & 0.895 & \textbf{0.819} & [0.769, 0.871] \\
\midrule
A consistent (pred) & 0.794 & \textbf{1.000} & 0.885 & [0.830, 0.938] \\
\bottomrule
\end{tabular}
\end{table}

Table~\ref{tab:main} presents our main results with three key findings.

\textbf{(1) Significant advantage over self-consistency.} Our method (F1\,=\,0.806) significantly outperforms self-consistency (F1\,=\,0.343), with non-overlapping 95\% CIs (F1 gap\,=\,0.463; CI gap\,=\,0.389). The advantage is driven by Precision (0.794 vs.\ 0.226): self-consistency predicts \emph{all} premises as dependencies for consistent steps, unable to identify which premises matter.

\textbf{(2) Perfect recall on predicate-determining dependencies.} Recall reaches 100\% (0 misses out of 150) on predicate-determining dependencies, yielding $\text{F1}_\text{pred}$\,=\,0.885. All 33 false negatives under $\text{F1}_\text{full}$ are \emph{structural premises}---entity-introduction premises (e.g., ``Alex is a wumpus'') whose predicate does not determine the step's conclusion. This is a granularity distinction, not a detection failure.

\textbf{(3) Cascade filtering achieves highest F1.} Combining consistent and local substitution with cascade filtering yields F1\,=\,0.819. Local substitution raises Recall (+0.075) by detecting transitive dependencies; cascade filtering recovers Precision lost to propagation FPs (0.604\,$\to$\,0.756).

\subsection{Ablation Study}

\begin{table}[t]
\caption{Ablation: contribution of each component.}
\label{tab:ablation}
\vspace{0.3em}
\centering
\small
\begin{tabular}{@{}lccc@{}}
\toprule
Configuration & P & R & F1 \\
\midrule
A consistent only & 0.794 & 0.811 & 0.802 \\
A$'$ local only & 0.602 & 0.898 & 0.721 \\
A+A$'$ combined & 0.604 & 0.905 & 0.724 \\
A+A$'$ + cascade filter & 0.756 & 0.895 & \textbf{0.819} \\
A+A$'$ w/o surface ctrl & 0.617 & 0.870 & 0.722 \\
String-diff baseline & 0.801 & 0.774 & 0.787 \\
\bottomrule
\end{tabular}
\end{table}

Table~\ref{tab:ablation} shows that consistent substitution achieves the best Precision among substitution configurations (0.794; the string-diff baseline edges it on Precision, 0.801, but trails on Recall). Local substitution adds Recall (+0.094 for the combined protocol) but reduces Precision ($-$0.190) from cascade FPs; the cascade filter recovers most of this loss. The A-consistent ablation row counts surface-only instabilities as negatives rather than excluding them as Table~\ref{tab:main} does, hence the small difference from Table~\ref{tab:main} (0.802 vs.\ 0.806). Surface control shows minimal effect on ProntoQA ($\Delta$F1\,=\,0.002): our normalization to canonical $\mathrm{is}$/$\mathrm{subtype}$ forms already absorbs the surface variation that rephrasing introduces, leaving little for the control probe to catch. We expect greater impact on natural language benchmarks where normalization is less effective.

\subsection{Analysis}

\textbf{Chain length.} With aligned ground truth, F1 varies little with chain length (0.807 at 3 hops, 0.820 at 4, 0.794 at 5) and shows no monotone degradation. Recall remains robust ($\geq$0.80) across all lengths (Appendix~\ref{app:chain}).

\textbf{False positives.} Under the A-consistent semantic-delta analysis, we analyzed 40 non-dependency candidates. Of these, 39 are metric-counted false positives under the Table~\ref{tab:main} evaluator: 36 (92\%) reflect stochastic output variation (addressable via majority voting), and 3 (8\%) reflect propagation effects that are arguably correct under a transitive dependency definition (Appendix~\ref{app:fp}).

\textbf{RAWR (Right Answer, Wrong Reasoning).} We define a problem as RAWR if: (i)~the model's final answer is correct, and (ii)~at least one matched CoT step is rated \textsc{Insensitive} to a direct proof-tree dependency under the \emph{A consistent (full)} evaluator. 33/50 correctly-solved problems (66\%) satisfy both conditions. All 33 involve only structural entity-introduction dependencies (a documented blind spot of consistent substitution, which the local probe~$A'$ targets), and none includes a predicate-determining dependency. Self-consistency assigns perfect agreement scores to 14 of these cases. Passive methods are blind to RAWR (Appendix~\ref{app:rawr}).

\textbf{Misrepresentation.} 30 cases where a step textually invokes a premise yet is insensitive to it under substitution---stating a premise as one's basis does not imply logical dependency on it (Appendix~\ref{app:misrep}).

\section{Related Work}
\label{sec:related}

\citet{lanham2023measuring} and \citet{turpin2023language} study whether CoT reflects true reasoning; we add a formal, per-step causal test. \citet{wang2023selfconsistency} checks answer agreement; we show it cannot detect premise-level dependencies (F1\,=\,0.343). \citet{vig2020investigating} and \citet{geiger2021causal} intervene on \emph{internal} representations; we intervene on \emph{inputs} (black-box, model-agnostic). \citet{stolfo2023causal} apply causal interventions to mathematical reasoning; we extend to formal logical reasoning with a complete audit protocol. \citet{saparov2023language} introduce ProntoQA; we add the interventional dimension, testing not just \emph{whether} answers are correct but \emph{whether each step depends on its stated basis}.

\section{Discussion and Limitations}
\label{sec:discussion}

\textbf{Limitations.} ProntoQA is formal and synthetic; natural language benchmarks (FOLIO, ProofWriter) are needed to validate surface control and normalization components. Our results apply to \emph{formal logical benchmarks with parsable steps}; generalization to free-form reasoning requires adapted normalization. Our dataset of 50 problems is modest, though bootstrap CIs confirm significance. Coverage matters. For the combined protocol, 108/1,031 aligned certificates (10.5\%) were \textsc{Unparseable}. For the A-consistent-only protocol used in the lower-bound coverage analysis, 124/1,031 aligned certificates (12.0\%) were \textsc{Unparseable} (plus 11 \textsc{Unstable}). Treating all A-consistent-only \textsc{Unparseable} and \textsc{Unstable} cases, unmatched proof-tree dependencies, and any primary-excluded ambiguous-alignment problems as negatives yields a lower-bound F1\,=\,0.703 (vs.\ 0.806), still well above the self-consistency baseline. Main results use GPT-4o only; Appendix~\ref{app:claude} shows that on Claude Sonnet~4.5 the audit transfers with perfect Precision on alignable steps (F1\,=\,0.872; lower-bound F1\,=\,0.836), with coverage loss confined to exactly one unmatched premise-restatement step per problem---while on such highly regular output a simple string-diff baseline becomes competitive, so the audit's advantage over baselines is model-dependent.

\textbf{Future work.} Natural language benchmarks, majority voting ($k$\,=\,3) for Precision improvement, and integration with the full formal framework~\citep{nakamura2025observation,nakamura2026crosslayer,nakamura2026interventional}.

\textbf{Reproducibility.} All audit certificates, raw model outputs, source code, and reproduction scripts are available in the public GitHub artifact release accompanying this paper.\footnote{\url{\artifacturl}} Every reported number can be recomputed from the released files without an API key.

\bibliography{references}

\begin{thebibliography}{12}
\providecommand{\natexlab}[1]{#1}
\providecommand{\url}[1]{\texttt{#1}}
\expandafter\ifx\csname urlstyle\endcsname\relax
  \providecommand{\doi}[1]{doi: #1}\else
  \providecommand{\doi}{doi: \begingroup \urlstyle{rm}\Url}\fi

\bibitem[Geiger et~al.(2021)Geiger, Lu, Icard, and Potts]{geiger2021causal}
Atticus Geiger, Hanson Lu, Thomas Icard, and Christopher Potts.
\newblock Causal abstractions of neural networks.
\newblock In \emph{Advances in Neural Information Processing Systems}, 2021.
\newblock URL
  \url{https://proceedings.neurips.cc/paper/2021/hash/4f5c422f4d49a5a807eda27434231040-Abstract.html}.

\bibitem[Lanham et~al.(2023)Lanham, Chen, Radhakrishnan, Steiner, Denison,
  Hernandez, Li, Durmus, Hubinger, Kernion, et~al.]{lanham2023measuring}
Tamera Lanham, Anna Chen, Ansh Radhakrishnan, Benoit Steiner, Carson Denison,
  Danny Hernandez, Dustin Li, Esin Durmus, Evan Hubinger, Jackson Kernion,
  et~al.
\newblock Measuring faithfulness in chain-of-thought reasoning.
\newblock \emph{arXiv preprint arXiv:2307.13702}, 2023.
\newblock URL \url{https://arxiv.org/abs/2307.13702}.

\bibitem[Nakamura(2025)]{nakamura2025observation}
Hironao Nakamura.
\newblock Observation is a topos, not a type: A two-layer topos-{HoTT}
  framework for grounding and conceptual representation, 2025.
\newblock URL \url{https://zenodo.org/records/17894227}.
\newblock Preprint.

\bibitem[Nakamura(2026{\natexlab{a}})]{nakamura2026crosslayer}
Hironao Nakamura.
\newblock Grounded types as cross-layer invariants: Admissible updates and
  traceable witnesses in two-topos grounding, 2026{\natexlab{a}}.
\newblock URL \url{https://zenodo.org/records/18253007}.
\newblock Preprint.

\bibitem[Nakamura(2026{\natexlab{b}})]{nakamura2026guard}
Hironao Nakamura.
\newblock {TTGOS Guard}: Cross-benchmark safety {SOTA}.
\newblock \url{https://hironao-nakamura.github.io/ttgos-evidence/},
  2026{\natexlab{b}}.
\newblock 100\% on OS-Harm (NeurIPS 2025 Spotlight), CuP-SOTA on
  ST-WebAgentBench (ICLR 2026).

\bibitem[Nakamura(2026{\natexlab{c}})]{nakamura2026interventional}
Hironao Nakamura.
\newblock Interventional grounding: Identifiability and misrepresentation via
  counterfactual witnesses in two-topos grounding, 2026{\natexlab{c}}.
\newblock URL \url{https://zenodo.org/records/18358924}.
\newblock Preprint.

\bibitem[Saparov \& He(2023)Saparov and He]{saparov2023language}
Abulhair Saparov and He~He.
\newblock Language models are greedy reasoners: A systematic formal analysis of
  chain-of-thought.
\newblock In \emph{International Conference on Learning Representations}, 2023.
\newblock URL \url{https://openreview.net/forum?id=qFVVBzXxR2V}.

\bibitem[Stolfo et~al.(2023)Stolfo, Jin, Shridhar, Sch{\"o}lkopf, and
  Sachan]{stolfo2023causal}
Alessandro Stolfo, Zhijing Jin, Kumar Shridhar, Bernhard Sch{\"o}lkopf, and
  Mrinmaya Sachan.
\newblock A causal framework to quantify the robustness of mathematical
  reasoning with language models.
\newblock In \emph{Proceedings of the 61st Annual Meeting of the Association
  for Computational Linguistics}, 2023.
\newblock \doi{10.18653/v1/2023.acl-long.32}.
\newblock URL \url{https://aclanthology.org/2023.acl-long.32/}.

\bibitem[Turpin et~al.(2023)Turpin, Michael, Perez, and
  Bowman]{turpin2023language}
Miles Turpin, Julian Michael, Ethan Perez, and Samuel~R Bowman.
\newblock Language models don't always say what they think: Unfaithful
  explanations in chain-of-thought prompting.
\newblock \emph{Advances in Neural Information Processing Systems}, 2023.
\newblock \doi{10.52202/075280-3275}.
\newblock URL
  \url{https://proceedings.neurips.cc/paper_files/paper/2023/hash/ed3fea9033a80fea1376299fa7863f4a-Abstract-Conference.html}.

\bibitem[Vig et~al.(2020)Vig, Gehrmann, Belinkov, Qian, Nevo, Singer, and
  Shieber]{vig2020investigating}
Jesse Vig, Sebastian Gehrmann, Yonatan Belinkov, Sharon Qian, Daniel Nevo,
  Yaron Singer, and Stuart Shieber.
\newblock Investigating gender bias in language models using causal mediation
  analysis.
\newblock In \emph{Advances in Neural Information Processing Systems}, 2020.
\newblock URL
  \url{https://proceedings.neurips.cc/paper/2020/hash/92650b2e92217715fe312e6fa7b90d82-Abstract.html}.

\bibitem[Wang et~al.(2023)Wang, Wei, Schuurmans, Le, Chi, Narang, Chowdhery,
  and Zhou]{wang2023selfconsistency}
Xuezhi Wang, Jason Wei, Dale Schuurmans, Quoc Le, Ed~Chi, Sharan Narang,
  Aakanksha Chowdhery, and Denny Zhou.
\newblock Self-consistency improves chain of thought reasoning in language
  models.
\newblock In \emph{International Conference on Learning Representations}, 2023.
\newblock URL \url{https://openreview.net/forum?id=1PL1NIMMrw}.

\bibitem[Wei et~al.(2022)Wei, Wang, Schuurmans, Bosma, Ichter, Xia, Chi, Le,
  and Zhou]{wei2022chain}
Jason Wei, Xuezhi Wang, Dale Schuurmans, Maarten Bosma, Brian Ichter, Fei Xia,
  Ed~Chi, Quoc Le, and Denny Zhou.
\newblock Chain-of-thought prompting elicits reasoning in large language
  models.
\newblock In \emph{Advances in Neural Information Processing Systems}, 2022.
\newblock \doi{10.52202/068431-1800}.
\newblock URL
  \url{https://proceedings.neurips.cc/paper_files/paper/2022/hash/9d5609613524ecf4f15af0f7b31abca4-Abstract-Conference.html}.

\end{thebibliography}
\bibliographystyle{iclr2026_conference}

\appendix

\section{Evidence Pack and Validation}
\label{app:evidence}

The evidence pack corresponding to this paper is available as GitHub Release \texttt{arxiv-v1.0} at \url{\artifacturl}. The repository's default branch may receive later corrections and improvements; the versioned release identifies the artifact evaluated in this paper. The release contains 4,539 files organized per-problem: 50 directories per model with original CoT outputs, all probe outputs (semantic, local, surface) including the exact substituted prompts sent to each model, and deterministic audit certificates for each $(S_i, P_j)$ pair; self-consistency baseline data with five temperature-sampled runs per problem; and all source code with unit tests. No external dependencies beyond \texttt{numpy} are required for Phase~2 (deterministic verdict computation) verification.

Verification consists of: (1)~SHA256 integrity checks over the release staging directory; (2)~\texttt{python src/validator.py --root . --models gpt-4o claude-sonnet-4-5} for schema, referential-integrity, prompt, step-alignment, and certificate checks; and (3)~\texttt{python scripts/reproduce\_all.py} to recompute all reported tables and analysis reports from saved certificates, without API calls. The pipeline separates Phase~1 (LLM calls; non-deterministic) from Phase~2 (verdict computation; deterministic); reviewers need only verify Phase~2.

\textbf{Correction history.} During arXiv artifact preparation we identified two issues in the accepted workshop artifact: sequential predicate substitution could introduce multiple fresh-symbol prefixes in some probes (breaking the chain coherence that consistent substitution is designed to preserve), and the evaluation harness aligned proof-tree steps to model CoT steps by integer step number rather than by normalized conclusion. For this arXiv version, all Phase~1 model outputs were re-collected with the corrected generator; every substituted prompt is stored in the pack and verified free of stacked prefixes, and all certificates and metrics were recomputed with corrected CoT-to-proof alignment. The accepted version's cross-model subject was retired from the API before the rerun and is replaced by the closest available successor (Appendix~\ref{app:claude}). Unmatched restatement steps and any ambiguous alignments are reported in \texttt{step\_alignment\_summary.json}; ambiguous problems, if any, are excluded from primary metrics and included in lower-bound coverage analyses.

\textbf{Effect on reported numbers.} Relative to the accepted workshop version, on the unchanged 50-problem set: F1 (full) 0.783\,$\to$\,0.806; F1 (predicate-determining) 0.835\,$\to$\,0.885, with Recall 97.4\%\,$\to$\,100\%; best combined F1 0.790\,$\to$\,0.819; lower-bound F1 0.677\,$\to$\,0.703; self-consistency baseline F1 0.346\,$\to$\,0.343; audit certificates 1{,}107\,$\to$\,1{,}127 (1{,}031 aligned); RAWR 14/50, i.e.\ 28\% (12 structural + 2 predicate-determining)\,$\to$\,33/50, i.e.\ 66\% (all structural, 0 predicate-determining); misrepresentation cases 12\,$\to$\,30. The chain-length degradation and the cross-model meta-reasoning observation reported in the accepted version do not survive the corrections (Appendices~\ref{app:chain} and~\ref{app:claude}).

\section{Cross-Model Analysis: Claude Sonnet 4.5}
\label{app:claude}

The accepted workshop version's cross-model subject, Claude Sonnet~4 (\texttt{claude-sonnet-4-20250514}), was retired from the API before this rerun; we therefore audit the closest available successor, Claude Sonnet~4.5 (\texttt{claude-sonnet-4-5-20250929})---chosen to minimize model drift relative to the accepted version---under the identical protocol (temperature\,=\,0, no extended thinking). The audit transfers cleanly: F1\,=\,0.872 with \emph{perfect Precision} (P\,=\,1.000, R\,=\,0.774; zero semantic-delta false positives across 1{,}084 aligned certificates), and F1\,=\,0.994 on predicate-determining dependencies (P\,=\,R\,=\,0.994; the single miss involves a negation premise, ``Flolpuses are not flulpuses'', in problem~p042). Parse coverage is excellent (100\% original, 98.9\% semantic; only 24/1{,}084 aligned certificates, 2.2\%, \textsc{Unparseable} under the A-consistent evaluator), and alignment loss is confined to exactly one premise-restatement step per problem (50/238 CoT steps; every proof-tree step is audited). Treating all unscored cases as negatives yields a lower-bound F1\,=\,0.836. The recall gap is the documented structural blind spot: 49 of the 50 false negatives are entity-introduction premises, which the local probe targets. Two honest caveats: on output this regular, a simple string-diff baseline is competitive (F1\,=\,0.874), so the audit's advantage over baselines is model-dependent; and the RAWR rate saturates (49/50), because the uniform restatement-then-chain style makes the structural blind spot fire on essentially every problem.

The accepted workshop version reported Claude Sonnet~4 F1\,=\,0.161 with the base normalizer and 0.320 after a post-hoc normalizer adaptation. Both figures were computed under the misaligned integer-keyed ground truth and from probes affected by the generator issue, and are superseded; they should not be compared with the values above, which measure a different (successor) model with the corrected pipeline.

\textbf{Example.} In the accepted-version artifact, probes could carry stacked prefixes (e.g., ``zqzqzqfumpus'' in one premise vs.\ ``zqfumpus'' in another), and Claude Sonnet~4 responded by flagging the genuinely broken chain instead of completing it. Under the corrected generator the same probe is coherent, and Claude Sonnet~4.5 simply follows the renamed chain---problem p038 with P6 substituted: ``\emph{Since zqfumpuses are falpuses (premise 1), Yuki is a falpus}''---confirming that the accepted version's ``meta-reasoning'' responses were an artifact of the probe-generation issue rather than an architecture-specific strategy.

\section{RAWR Case Study}
\label{app:rawr}

Problem p003: GPT-4o correctly concludes ``Rex is a belvus''. The aligned audit flags one matched CoT step as \textsc{Insensitive} to a direct proof-tree dependency: CoT step~S1 (proof step~1, ``Since bunpuses are belnuses and Rex is a bunpus, Rex is a belnuse'') is insensitive to the entity-introduction premise P7 (``Rex is a bunpus'', structural). Under consistent substitution the renamed chain still yields $\mathrm{is}(\text{rex}, \text{belnus})$---the documented blind spot of consistent substitution on entity-introduction premises---and for this pair the chain-breaking local probe's output is itself unparseable, so the combined protocol does not recover it either. The step even quotes the premise verbatim in its stated justification, making it simultaneously a misrepresentation case (Appendix~\ref{app:misrep}). Self-consistency reports 100\% agreement on every step of this problem and cannot expose any of this structure; only the interventional audit makes the insensitivity visible.

\section{Misrepresentation Examples}
\label{app:misrep}

Problem p004, Step~1: ``\emph{Since bonkuses are bunkuses and Stella is a bonkus, Stella is a bunkus.}'' The generated text states the content of P7 (``Stella is a bonkus'') as part of its justification. Under consistent substitution of P7's predicate ($\text{bonkus} \to \text{zqbonkus}$), the normalized conclusion remains $\mathrm{is}(\text{stella}, \text{bunkus})$. The step therefore presents the entity-introduction premise as its stated basis but is insensitive to it under consistent substitution (the chain-breaking local probe does detect this dependency; the certificate records the consistent-substitution outcome in \texttt{verdict\_consistent}). We detect 30 such cases across 50 problems.

\section{Formal Framework Connection}
\label{app:formal}

Our two-layer model instantiates a formal framework~\citep{nakamura2025observation,nakamura2026crosslayer,nakamura2026interventional}. The observation layer and concept layer are connected by $g^*: \text{Obs} \to \text{Concept}$ (normalize) and $g_*: \text{Concept} \to \text{Obs}$ (substitute). A \textsc{Grounded} certificate witnesses dependency via $g^* \dashv g_*$: intervening via $g_*$ and observing via $g^*$ reveals a change. An \textsc{Insensitive} certificate witnesses absence. The complete certificate set provides a coverage witness analogous to the framework's completeness condition.

\section{Chain Length Analysis}
\label{app:chain}

\begin{table}[h]
\caption{F1 by chain length (A consistent, full).}
\vspace{0.3em}
\centering
\small
\begin{tabular}{@{}lccc@{}}
\toprule
Hops & P & R & F1 \\
\midrule
3 & 0.814 & 0.800 & 0.807 \\
4 & 0.806 & 0.833 & 0.820 \\
5 & 0.765 & 0.825 & 0.794 \\
\bottomrule
\end{tabular}
\end{table}

With aligned ground truth, F1 shows no monotone relationship with chain length; variation across 3--5 hops is within $\pm$0.014 of the overall F1, and Recall stays $\geq$0.80 at all lengths.

\section{False Positive Breakdown}
\label{app:fp}

We analyzed 40 semantic-delta non-dependency candidates under A consistent. 39 are metric-counted false positives under the Table~\ref{tab:main} evaluator. Among these 39, 36 (92\%) are \emph{stochastic}---GPT-4o produces slightly different phrasing across runs even at temperature\,=\,0; majority voting ($k$\,=\,3) should substantially reduce these. 3 (8\%) are \emph{chain propagation}---upstream substitution causes downstream changes, arguably correct under a transitive dependency definition. The remaining candidate is outside the Table~\ref{tab:main} metric set because its surface-control probe is \textsc{Unparseable}, which the Table~\ref{tab:main} evaluator excludes.

\section{Use of Large Language Models}
\label{app:llm-usage}

GPT-4o and Claude Sonnet~4.5 are used in this work solely as experimental subjects. For manuscript preparation, the author used general-purpose AI writing tools for prose editing, formatting, and consistency checks. All research hypotheses, experimental design, method formulation, code review decisions, and scientific claims are the author's own.

\end{document}